\pdfoutput=1
\documentclass[11pt,a4paper]{article}
\usepackage[hyperref]{naaclhlt2019}
\usepackage{times}
\usepackage{amsmath}
\usepackage{latexsym}
\usepackage{placeins}
\usepackage{float}
\usepackage{url}
\usepackage{subcaption}
\usepackage{graphicx}

\aclfinalcopy 

\title{Jointly Measuring Diversity and Quality \\ in Text Generation Models}

\author{Ehsan Montahaei\thanks{\quad These authors contributed equally to this work.} \\
  Sharif University of \\ Technology / Tehran, Iran \\
  \small{\tt ehsan.montahaei@gmail.com} \\\And
  Danial Alihosseini\footnotemark[1] \\
  Sharif University of \\ Technology / Tehran, Iran \\
  \small{\tt dalihosseini@ce.sharif.edu} \\\And
  Mahdieh Soleymani Baghshah \\
  Sharif University of \\ Technology / Tehran, Iran \\
  \small{\tt soleymani@sharif.edu} \\}

\date{}

\begin{document}
\maketitle
\begin{abstract}
Text generation is an important Natural Language Processing task with various applications. Although several metrics have already been introduced to evaluate the text generation methods, each of them has its own shortcomings. The most widely used metrics such as BLEU only consider the quality of generated sentences and neglect their diversity.
For example, repeatedly generation of only one high quality sentence would result in a high BLEU score. On the other hand, the more recent metric introduced to evaluate the diversity of generated texts known as Self-BLEU ignores the quality of generated texts. In this paper, we propose metrics to evaluate both the quality and diversity simultaneously by approximating the distance of the learned generative model and the real data distribution. For this purpose, we first introduce a metric that approximates this distance using n-gram based measures. Then, a feature-based measure which is based on a recent highly deep model trained on a large text corpus called BERT is introduced. Finally, for oracle training mode in which the generatorʼs density can also be calculated, we propose to use the distance measures between the corresponding explicit distributions. 
Eventually, the most popular and recent text generation models are evaluated using both the existing and the proposed metrics and the preferences of the proposed metrics are determined.
\end{abstract}

\section{Introduction}
Generative models and especially Generative Adversarial Networks (GANs) have been received much attention in the last few years. However, the evaluation of generated samples by these models is challenging. Although some studies have recently focused on introducing measures like Inception Score and Fr\'echet Inception Distance (FID) to compare results of different GAN models for image generation, there is not a study to propose proper metrics for evaluation of text generation models. 
In the last few years, many GAN-based text generation models \cite{seqgan,rankgan,maligan,leakgan,textgan} have been proposed. However, measuring the performance of these models in the corresponding papers is not comprehensive.
GANs suffer from the mode collapse problem 
\cite{UnrolledGAN} and the GAN-based text generation models may just produce a highly limited set of sentences and therefore just considering the quality of these generated sentences for comparison is not comprehensive.

On the other hand, there are measures like Self-BLEU \cite{textgan} for evaluating the diversity of generated sentences, but they can not consider the quality of samples at all. Besides, designing an experiment of evaluating diversity by humans is not straightforward and thus it's necessary to have a jointly quality-diversity measuring metric.
 
In this paper, we intend to propose metrics sensitive to both quality and diversity simultaneously, assigning low scores not only to models generating low-quality samples but also to the ones with low-diversity samples (including the mode collapsed models). 
To this end, we first propose the MS-Jaccard as an n-gram based measure that considers the quality and diversity of generated samples simultaneously. It attempts to find the similarity of the set of generated samples by a model and the set of real (or test) samples. 
Then, a feature-based measure is proposed to compare the real data distribution and the generative model distribution in the feature space. Indeed, by borrowing the idea of FID \cite{heusel2017gans} that is a popular feature-based evaluation metric in image generation tasks and advent of a recent highly deep model named BERT \cite{devlin2018bert} as a reference feature extractor for natural language texts, a metric is proposed for evaluation of natural language generation. Finally, appropriate divergences between the oracle distribution and the (learned) model distribution is introduced
for when the probabilistic oracle is considered as synthetic data distribution (and thus the target distribution is available for evaluation). 

\section{Text Generation Models}
The neural models on text generation first used LSTMs and trained them by the Maximum Likelihood Estimation (MLE) via teacher forcing \cite{lstm}. These models suffer from the exposure bias problem which is due to the train-test discrepancy. Although some solutions such as scheduled sampling were introduced to overcome the exposure bias problem, it has been shown that they are incompatible with the language nature \cite{sched-samp, ss-dis}.
By introducing GANs \cite{gan} as successful image generation models, it has gained much attention to propose GAN-based text generation models. However, the discrete nature of text needs the generator with discrete outputs that makes passing the gradient from the discriminator to the generator difficult.
SeqGAN \cite{seqgan} alleviates this difficulty by a gradient policy approach using a REINFORCE-like method to train the generator as a stochastic policy.
This method has some difficulties such as reward sparsity and high variance for large action spaces. 
Subsequent methods try to pass more informative signal from the discriminator to the generator. RankGAN\cite{rankgan} trains the discriminator as a ranker which assigns a higher score to the more realistic sequences (in comparison with other sentences in the current batch). 
LeakGAN \cite{leakgan} takes advantage of the feudal networks and considers the discriminator as a manager and the generator as a worker while the feature layer of the discriminator is fed to the generator as leaked information.
MaliGAN \cite{maligan} attempts to redefine the generator's objective. It minimizes KL divergence between the generator and the real distribution which is obtained by the discriminator in the optimality assumption of the discriminator. This new objective leads to an importance sampling procedure.
TextGAN \cite{textgan} also applies a new objective for the generator. It tries to push the generator focus from the last layer of the discriminator to its last feature layer. Real data and generator samples will each have some distribution in the feature layer of the discriminator. The generator's objective is to make them closer by Maximum Mean Discrepancy (MMD) metric.

\section{Metrics}
In this section, we first indicate the main difficulties of the existing measures for evaluation of text generation models. Then, we introduce metrics that evaluate the capability of the models in generating both right sentences and various ones. The proposed metrics (that are all symmetric) jointly specify to what extent probable sentences in real data are likely in the generative model and also the probable sentences in the model are likely in the real data.

\subsection{Shortcomings of the existing metrics}
In this section, shortcomings of the metrics that either evaluate the quality or the diversity of generated samples are presented. Moreover, a recent attempt to simultaneously considering these metrics is introduced.
\subsubsection{Quality metrics}
\textbf{BLEU}: It is the most widely used metric for text generation. Originally BLEU \cite{bleu} is a metric to evaluate the quality of machine-translated text.
In unconditional text generation, all sentences in the test set are considered as the reference set and generated sentences are evaluated by computing their average BLEU score on this reference set. 
In conditional text generation tasks like machine translation which include a limited reference set (for each condition), computing the similarity of the generated text and the reference set may be sensible. However, the reference set for the unconditional text generation task is whole available sentences and measures like BLEU just consider the validity of generated sentences without measuring what proportion of the reference sentences can be covered by the text generation model.
On the other hand, GAN-based text generation models may generate a highly limited set of sentences and sacrifice the diversity (due to the mode collapse problem). Therefore, evaluating these models using BLEU score just shows the validity of their outputs without considering their coverage.

\textbf{Oracle-NLL}: It was introduced by SeqGAN \cite{seqgan} and is based on assuming a synthetic oracle distribution. It considers a random distribution as the real distribution (or the oracle) and the training dataset is prepared by sampling from this distribution. The score is defined to be the Negative Log Likelihood (NLL) of the generated samples from the trained model in the oracle distribution. In this measure, the coverage is again neglected and a model that generates only one high quality sentence can reach high performance. 

\subsubsection{Diversity metric}
As mentioned above, BLUE and Oracle-NLL just consider the quality of the generated samples and ignore their diversity. Below, we introduce two metrics measuring the diversity. However, these metrics evaluate only diversity and don't consider the quality of samples at all.

\textbf{Self-BLEU}: In \cite{zhu2018texygen}, Self-BLEU was introduced to evaluate just variety of sentences. It measures BLEU score for each generated sentence by considering other generated sentences as reference. By averaging these BLEU scores (obtained for generated sentences), a metric that is called Self-BLEU is achieved where its lower values shows more diversity.

\textbf{Entropy}: On the other side, we can use the entropy of probabilistic generative model to measure the diversity where the lower values show lower diversity. As the direct calculation of the entropy is not feasible, a Monte-Carlo estimation of it can be used.

\subsubsection{Quality and diversity} \label{quality_diversity}
Recently \cite{Caccia} mentioned the flaws of only evaluating the quality and found that MLE outperforms the GAN variants for text generation since it dominates GANs in the quality-diversity space.
\cite{Caccia} uses the quality-diversity spectrum obtained by changing the temperature parameter that controls entropy of the models' conditional distributions.
However, it does not provide a measure to assess both the quality and the diversity without needing to inspect the whole quality-diversity spectrum.

\textbf{Likelihood}: Although the likelihood of a generative model on real (test) data evaluates the ability of the model in generating the test samples, it doesn't measure the quality of the whole set of generated texts by the model. In fact, a model with a low NLL value on test data (or equivalently a model in which the likelihood of the test data is high) may also assign high probability to many other sentences that are not valid or qualified. 
Specifically for sequence models, the likelihood doesn't assess the free-running mode of models. To be more detailed, most of the probabilistic sequence models, decompose the joint distribution to conditional distributions using the chain rule. These conditional distributions are the probability of each token conditioned on the prior tokens. Thus, in the likelihood evaluation, each of token's probability is conditioned on a prefix that is a real sequence itself and the likelihood is not assessed on the previously generated tokens of the model during evaluation (it is similar to the exposure bias problem of MLE for sequence generation).

Moreover, measuring a model by its likelihood score has another problem. When a model misses one mode of a multi-modal distribution, its score decreases severely; so it is an unfair metric for comparing MLE method with other methods because MLE method uses likelihood as its objective and has mean seeking behavior \cite{nips_tutorial}.


\subsection{Proposed metrics}
In this section, we propose metrics that simultaneously considers the quality and the diversity of the generated samples. To this end, we compare the real distribution of texts with the obtained distribution by the text generation model. 
\subsubsection{MS-Jaccard}
We first propose a metric that finds the similarity of the generative model and the real distribution by comparing text samples generated by them. To this end, n-grams of generated samples and those of real samples are considered as two multi-sets (that also preserve repetition of n-grams) and the similarity of the resulted multi-sets is computed.
In simple words, the MS-Jaccard focuses on the similarity of the n-grams frequencies in the two sets and inspired by the well-known Jaccard Index which determines the similarity of two sets as the ratio of the cardinality of their intersection to that of their union.

To define it formally, let $S_1$ and $S_2$ be two sets of sentences, $G_n$ be the set of n-grams in $S_1 \cup S_2$, and $C_n(g, S)$ be the normalized counts of the n-gram $g$ in the set $S$. The  similarity between n-grams of two sets $S_1$ and $S_2$ is defined as:
\begin{equation}
{score_n = \frac{
\sum_{g \in G_n} min\{C_n(g,S_1), C_n(g, S_2)\}}
{\sum_{g \in G_n} max\{C_n(g,S_1), C_n(g, S_2)\}}}.
\end{equation}
The geometric mean of the $\{{score_n\}}_{n=1}^{N}$ will be the MS-Jaccard score called MS-Jaccard-$N$ where the $N$ is the maximum length of $n$-grams. It is worth noting that the frequencies of the n-grams in each set is normalized with respect to the total number of sentences in the set (to avoid diminishing the score when the size of only one of these sets grows).
Thus, the $C_n(g, S)$ will denotes the average frequency per sentence for n-gram $g$ in the set $S$. If the generated sentences won't have diversity or quality, the n-gram distribution of generated texts will be different from that of the real texts and causing to decrease the MS-Jaccard score consequently.
As it is obvious, the MS-Jaccard is a similarity measure and so its higher value will be better.

\subsubsection{Fr\'echet BERT Distance (FBD)}
One popular metric for evaluation of image generation models is FID introduced in \cite{heusel2017gans}. Each of real and generated images in a feature space (found by Inception network) is modeled by a Gaussian distribution, and the FID is defined as the Fr\'echet distance between these two Gaussian distributions. We want to introduce a similar measure for the text generation task. To this end, we utilize BERT \cite{devlin2018bert} that provides a proper feature space for texts. We use Fr\'echet distance in BERT's feature space as a metric that considers quality and variety of generated sentences, and name it Fr\'echet BERT Distance (FBD). There is a set of pooled features (for classification task) in the BERT network that has a constant size for different input sequence lengths; we used these features for FBD.
The Fr\'echet distance is also known as Wasserstein-2 divergence, and this distance between two Gaussian distribution is as follows:
\begin{equation}
\sqrt{
{||m_1-m_2||}_2^2 + Tr( C_1 + C_2 -2(C_1C_2)^{1/2})},
\end{equation}
where $m_i$ and $C_i$ show the mean vector and the covariance matrix of these Gaussians respectively. It should be noted as the FBD is a distance measure, its lower values will be better.

\subsubsection{Oracle Based Evaluation} \label{measure:oracle}
In Oracle-NLL evaluation introduced in \cite{seqgan}, the measured distance is Kullback--Leibler (KL) divergence of the generative model and the oracle which ignores the variety of generated sentences. On the other hand, the inverse KL (that is relevant to the likelihood of real data in the text generation model) can not guarantee the quality of generated samples by the model. We propose measuring the distance of the probabilistic oracle distribution $P$ (that generates real data) and the probabilistic generative model $Q$ by a symmetric distance as an evaluation metric. A wide range of distances can be utilized for this purpose. One symmetric distance is Bhattacharyya that can be estimated by the Monte-Carlo as below:
\begin{equation}
    \label{bhattacharyya}
    \begin{aligned}
    B(P,Q) & =\\
    \frac{-1}{2} \Big(
        &\ln \frac{1}{N}\sum_{i=0}^N \sqrt{\frac{q(x_i)}{p(x_i)}}
        +\ln \frac{1}{M}\sum_{j=0}^M     \sqrt{\frac{p(x_j)}{q(x_j)}}
    \Big)
    ,
   \end{aligned}
\end{equation}
where $\{x_i\}$ and $\{x_j\}$ are sets of samples from $P$ and $Q$ distributions respectively. Similar to the FBD, Bhattacharyya is also a distance measure and thus its lower values are better.

\section{Evaluation}
In this section, we first conduct some experiments to evaluate text generation models using the existing and the proposed measures. Then, we discuss about the appropriateness of the proposed metrics.
\subsection{Datasets}
We evaluate the models on COCO image captions \cite{coco_dataset}, EMNLP2017 WMT News \cite{emnlp_dataset}, and IMDB \cite{imdb_dataset} as the popular datasets for text generation. In addition to these datasets, similar to \cite{seqgan, rankgan, leakgan}, we also consider a synthetic oracle produced by a probabilistic text generator that is a random initialized LSTM as a synthetic dataset.
The description of the datasets is as follows:
\begin{itemize}
    \item COCO Captions \cite{coco_dataset}: It is a collection of image captions containing around 600,000 captions. Sentences having between 5 and 25 words are selected (resulting in 524,225 sentences) where 5,328 is the vocab size of the resulted dataset. Among the resulted dataset, 40,000 samples are used for training, 20,000 samples for validation, and 20,000 for test.
    \item EMNLP2017 WMT News \cite{emnlp_dataset}: It is a collection of news texts for the machine translations task \footnote{\href{http://statmt.org/wmt17/translation-task.html}{http://statmt.org/wmt17/translation-task.html}}. Among a version of this dataset for English corpus containing 500,000 sentences, \textcolor{black}{sentences having more than 3 words with less than 150 frequency (these words are replaced with UNK) were dropped and sentences that have between 20 and 40 words selected}. The vocab size of the resulted dataset is 6,148. Among this dataset, 40,000 samples are used for training, 20,000 samples for validation, and 20,000 for test.
    \item IMDB Movie Reviews \cite{imdb_dataset}: It is a collection of IMDB movie reviews for the sentiment analysis task, containing 25,000 labeled and 50,000 unlabeled ones. We have selected the first two sentences of each review and replace words with less that 50 times frequency with UNK and keep sentences from length 5 to 40 with less than 5 UNKs. The final dataset is subsampled to have 20,000 sentences for training data, 10,000 for validation, and 10,000 for test data leading to vocab size of 5,810.
    \item Oracle synthetic dataset \cite{seqgan}: A randomly initialized LSTM generator as a real distribution used in oracle training mode; the network implementation is borrowed from the SeqGAN released code\footnote{\href{https://github.com/LantaoYu/SeqGAN/blob/master/target_lstm.py}{https://github.com/LantaoYu/SeqGAN/}}. This network's hidden size is 32 and its embedding size is 3,200. Moreover, the vocab size is 5,000 and the length of sequences is 20. The dataset of 100,000 samples are generated according to the above model. Among this dataset, 50,000 samples are used for training, 25,000 for validation, and 25,000 for test.
\end{itemize}
\begin{table*}[!htb]
\centering
\caption{Performance of models (using different measures) on \textit{COCO Captions} dataset. 
MSJ, BL, and SBL denote MS-Jaccard, BLEU, and Self-BLEU respectively.}

\small\tabcolsep=0.07cm
\begin{tabular}{||c||c|c|c c c c|c c c c|c c c c||}\hline\hline Method	& NLL	& FBD	& MSJ2	& MSJ3	& MSJ4	& MSJ5	& BL2	& BL3	& BL4	& BL5	& SBL2	& SBL3	& SBL4	& SBL5\\
\hline\hline
Real Data	& -	& $0.460$	& $0.760$	& $0.585$	& $0.430$	& $0.306$	& $0.926$	& $0.794$	& $0.622$	& $0.454$	& $0.864$	& $0.685$	& $0.489$	& $0.329$ \\
\hline
MLE	& $\mathbf{38.416}$	& $1.971$	& $0.655$	& $0.473$	& $0.322$	& $0.210$	& $0.891$	& $0.715$	& $0.507$	& $0.334$	& $\mathbf{0.849}$	& $\mathbf{0.644}$	& $\mathbf{0.425}$	& $\mathbf{0.268}$ \\
\hline
SeqGAN	& $55.610$	& $4.590$	& $0.301$	& $0.229$	& $0.164$	& $0.111$	& $0.904$	& $0.771$	& $\mathbf{0.578}$	& $\mathbf{0.380}$	& $0.941$	& $0.842$	& $0.700$	& $0.545$ \\
\hline
MaliGAN	& $39.916$	& $\mathbf{1.474}$	& $\mathbf{0.671}$	& $\mathbf{0.495}$	& $\mathbf{0.345}$	& $\mathbf{0.231}$	& $0.901$	& $0.736$	& $0.536$	& $0.361$	& $0.859$	& $0.662$	& $0.451$	& $0.288$ \\
\hline
RankGAN	& $48.816$	& $3.574$	& $0.440$	& $0.323$	& $0.224$	& $0.147$	& $\mathbf{0.927}$	& $\mathbf{0.782}$	& $0.569$	& $0.376$	& $0.913$	& $0.774$	& $0.583$	& $0.402$ \\
\hline
\hline\end{tabular}\small 
\label{table:COCO Captions}
 \end{table*}

\begin{table*}[!htb]
\centering

\caption{Performance of models (using different measures) on \textit{EMNLP2017 WMT News} dataset. MSJ, BL, and SBL denote MS-Jaccard, BLEU, and Self-BLEU respectively.}

\small\tabcolsep=0.07cm
\begin{tabular}{||c||c|c|c c c c|c c c c|c c c c||}\hline\hline Method	& NLL	& FBD	& MSJ2	& MSJ3	& MSJ4	& MSJ5	& BL2	& BL3	& BL4	& BL5	& SBL2	& SBL3	& SBL4	& SBL5\\
\hline\hline
Real Data	& -	& $0.905$	& $0.691$	& $0.432$	& $0.243$	& $0.129$	& $0.886$	& $0.644$	& $0.380$	& $0.198$	& $0.797$	& $0.512$	& $0.261$	& $0.133$ \\
\hline
MLE	& $\mathbf{143.246}$	& $\mathbf{4.827}$	& $\mathbf{0.585}$	& $\mathbf{0.334}$	& $\mathbf{0.164}$	& $\mathbf{0.071}$	& $0.837$	& $0.542$	& $0.264$	& $0.125$	& $\mathbf{0.777}$	& $\mathbf{0.452}$	& $\mathbf{0.196}$	& $\mathbf{0.095}$ \\
\hline
SeqGAN	& $195.867$	& $5.955$	& $0.231$	& $0.138$	& $0.071$	& $0.031$	& $0.476$	& $0.358$	& $0.200$	& $0.105$	& $0.906$	& $0.729$	& $0.507$	& $0.324$ \\
\hline
MaliGAN	& $163.931$	& $5.690$	& $0.405$	& $0.249$	& $0.132$	& $0.061$	& $\mathbf{0.856}$	& $\mathbf{0.595}$	& $\mathbf{0.314}$	& $\mathbf{0.141}$	& $0.847$	& $0.591$	& $0.328$	& $0.155$ \\
\hline
RankGAN	& $177.346$	& $5.104$	& $0.261$	& $0.156$	& $0.081$	& $0.036$	& $0.461$	& $0.326$	& $0.183$	& $0.097$	& $0.841$	& $0.605$	& $0.371$	& $0.224$ \\
\hline
\hline\end{tabular}\normalsize .
\label{table:EMNLP2017 WMT News}

 \end{table*}

\begin{table*}[!htb]
\centering
\caption{Performance of models (using different measures) on \textit{IMDB Movie Reviews} dataset. MSJ, BL, and SBL denote MS-Jaccard, BLEU, and Self-BLEU respectively.}

\small\tabcolsep=0.07cm
\begin{tabular}{||c||c|c|c c c c|c c c c|c c c c||}\hline\hline Method	& NLL	& FBD	& MSJ2	& MSJ3	& MSJ4	& MSJ5	& BL2	& BL3	& BL4	& BL5	& SBL2	& SBL3	& SBL4	& SBL5\\
\hline\hline
Real Data	& -	& $0.683$	& $0.696$	& $0.469$	& $0.296$	& $0.181$	& $0.889$	& $0.691$	& $0.468$	& $0.286$	& $0.853$	& $0.629$	& $0.405$	& $0.241$ \\
\hline
MLE	& $\mathbf{125.223}$	& $\mathbf{3.538}$	& $\mathbf{0.601}$	& $\mathbf{0.375}$	& $\mathbf{0.214}$	& $\mathbf{0.115}$	& $0.860$	& $0.620$	& $0.368$	& $0.198$	& $\mathbf{0.844}$	& $\mathbf{0.593}$	& $\mathbf{0.342}$	& $\mathbf{0.179}$ \\
\hline
SeqGAN	& $150.213$	& $4.587$	& $0.377$	& $0.247$	& $0.147$	& $0.082$	& $\mathbf{0.903}$	& $\mathbf{0.695}$	& $\mathbf{0.434}$	& $0.226$	& $0.924$	& $0.763$	& $0.552$	& $0.345$ \\
\hline
MaliGAN	& $141.558$	& $4.482$	& $0.446$	& $0.294$	& $0.178$	& $0.103$	& $0.878$	& $0.662$	& $0.424$	& $\mathbf{0.233}$	& $0.889$	& $0.695$	& $0.480$	& $0.290$ \\
\hline
RankGAN	& $151.828$	& $3.958$	& $0.354$	& $0.227$	& $0.132$	& $0.070$	& $0.900$	& $0.693$	& $0.432$	& $0.228$	& $0.909$	& $0.739$	& $0.527$	& $0.331$ \\
\hline
\hline\end{tabular}\normalsize 
\label{table:IMDB Movie Reviews}
 \end{table*}

\begin{table}[!htb]
\centering
\caption{Performance of models (using different measures) on \textit{Oracle} dataset.}\label{table:Oracle}

\small\tabcolsep=0.07cm
\begin{tabular}{||c||c c|c||}\hline\hline Method	& NLL	& Oracle-NLL	& Bhattacharyya\\
\hline\hline
MLE	& $\mathbf{141.948}$	& $167.014$	& $\mathbf{7.105}$ \\
\hline
SeqGAN	& $155.353$	& $\mathbf{163.179}$	& $10.076$ \\
\hline
MaliGAN	& $146.260$	& $168.054$	& $8.503$ \\
\hline
RankGAN	& $160.424$	& $166.774$	& $12.127$ \\
\hline
\hline\end{tabular}\normalsize 
 \end{table}

\subsection{Experimental Setup}
\subsubsection{Text Generation Models}
As the recent methods for text generation, we evaluate SeqGAN \cite{seqgan}, RankGAN \cite{rankgan}, and MaliGAN \cite{maligan}. 
We also consider vanilla Maximum Likelihood Estimation (MLE) language model using LSTM as the baseline method. We used the implementation of the above methods in the Texygen platform \cite{zhu2018texygen} and train them in this framework\footnote{\href{https://github.com/geek-ai/Texygen}{https://github.com/geek-ai/Texygen}}. The models were trained on the similar dataset existing in their released code but collected from the original sites reported in corresponding reference papers.

In order to have a fair comparison, all settings of the models (e.g., same hidden) were kept the same as the Texygen framework. 
Since setting a fixed number of epochs for terminating training of different methods does not seem such reasonable and resulting in unfair scores, we targeted multiple training termination criteria. In the real-world datasets training, the training termination of the GANs were based on obtaining the best BLEU4 on validation data in addition to setting a max number of iterations for all the models. Besides, the training termination of MLE is based the NLL on the validation data while also setting a max number of iterations as above. In the oracle training mode, the termination were done based on both Oracle-NLL on the validation set and again on a max number of iterations for all models.

\subsubsection{Metrics}

Among the existing measures, BLEU2 upto BLEU5 (evaluating only quality), Self-BLUE2 upto Self-BLEU5 (evaluating only diversity), and NLL that shows the negative log likelihood of the model on test data are utilized for real datasets. Moreover, due to the low performance of the Python NLTK \cite{nltk_book} BLEU library\footnote{\href{https://www.nltk.org/_modules/nltk/translate/bleu_score.html\#sentence_bleu}{https://www.nltk.org/\_modules/nltk/}}
when needing to evaluate multiple sentences with a fixed reference set, we have re-implemented it to achieve parallel computation and high performance \footnote{\href{https://github.com/Danial-Alh/FastBLEU}{https://github.com/Danial-Alh/FastBLEU}}.

Among the proposed measures, MS-Jaccard2 upto MS-Jaccard5 and FBD are assayed on real-world datasets. For synthetic oracle, NLL and Oracle-NLL as the existing measures and the proposed measure for comparing distributions, i.e. Bhattacharyya,  are evaluated. It should be noted that, in order to make the metric's directions the same (i.e. their lower values show better performance), the $1-$MS-Jaccard, $1-$BLEU and $-1 \times$Entropy is used in some plots.
\subsection{Results}
Results of different methods on COCO Captions, EMNLP2017 WMT News, and IMDB datasets as real-world datasets are shown in Tables \ref{table:COCO Captions}, \ref{table:EMNLP2017 WMT News}, and \ref{table:IMDB Movie Reviews}, respectively. To provide a target, we have also shown metrics for training data themselves and called the method as \textit{Real} (indeed training data is considered as the generated data by Real and the measures are computed on them). These tables show that MLE has the best performance according to the proposed measures considering both quality and diversity of samples. In fact, GAN-based methods can not generally achieve good performance according to the proposed measures. This result is consistent with the reported results in \cite{Caccia} that compares GANs and MLE for text generation.

Table \ref{table:Oracle} shows results of different methods on synthetic oracle dataset and MLE again shows the best results according to the proposed metric (that approximates the distance of the real distribution and the generative model distribution).  

As mentioned in Section \ref{quality_diversity} about \cite{Caccia}, the whole spectrum of quality-diversity is considered for evaluation of Natural Language Generation (NLG) methods. In fact, in \cite{Caccia}, the temperature sweep is utilized to robustly evaluate text generation methods. 
More precisely, the generator’s conditional distribution $G(x_{t}|x_{1:t-1})$ is defined as $Softmax(o_t/T)$ where $o_t$ denotes the logit at time $t$. Decreasing $T$ below $1.0$ will decrease  the  entropy  of conditional probability and thus reduce the probability of generating low quality samples. On the other hand, increasing this temperature above $1.0$ will upraise the entropy of the conditional distribution and thus improve the diversity of the generated samples \cite{Caccia}.

\begin{figure*}[!thb]
    \centering
    \begin{subfigure}[b]{0.40\textwidth}
        \centering
        \includegraphics[width=0.93\textwidth]{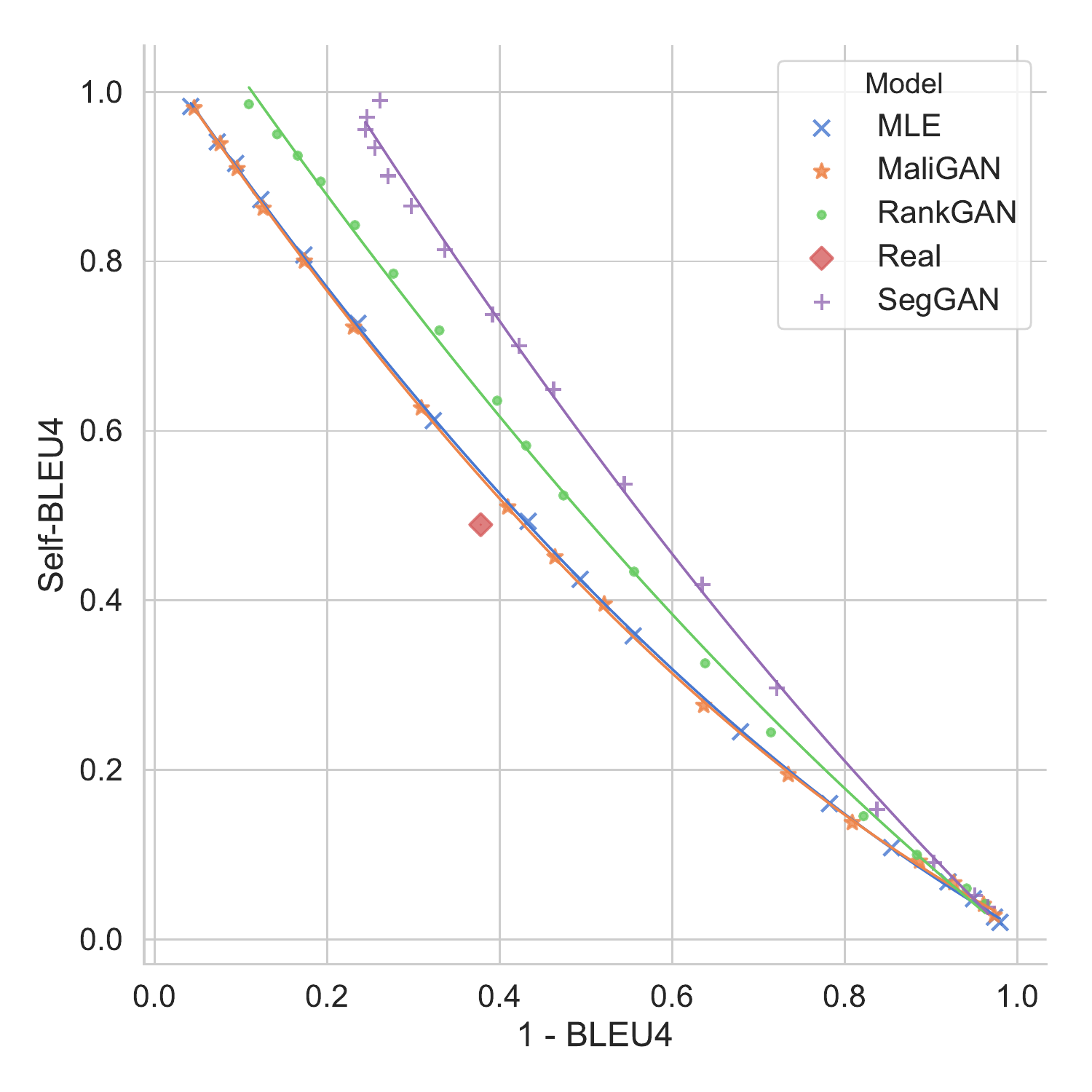}
    	\caption{\textit{COCO Captions} dataset}
    	\label{M2:coco}
    \end{subfigure}
    \quad
    \begin{subfigure}[b]{0.40\textwidth}
        \centering
    	\includegraphics[width=0.93\textwidth]{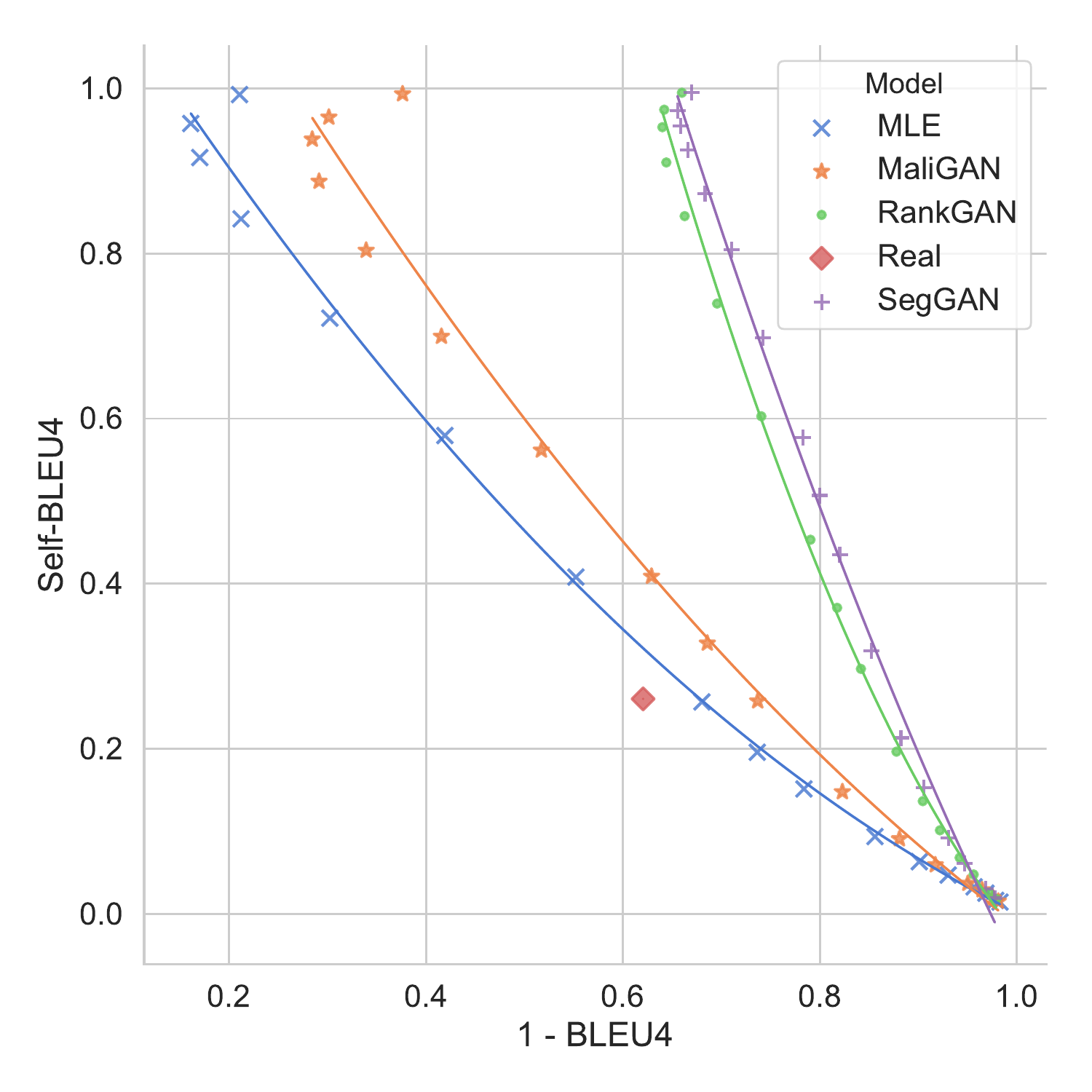}
    	\caption{\textit{EMNLP2017 WMT News} dataset}
    	\label{M2:emnlp}
    \end{subfigure}
    \vskip\baselineskip
    \begin{subfigure}[b]{0.40\textwidth}
    	\centering
    	\includegraphics[width=0.93\textwidth]{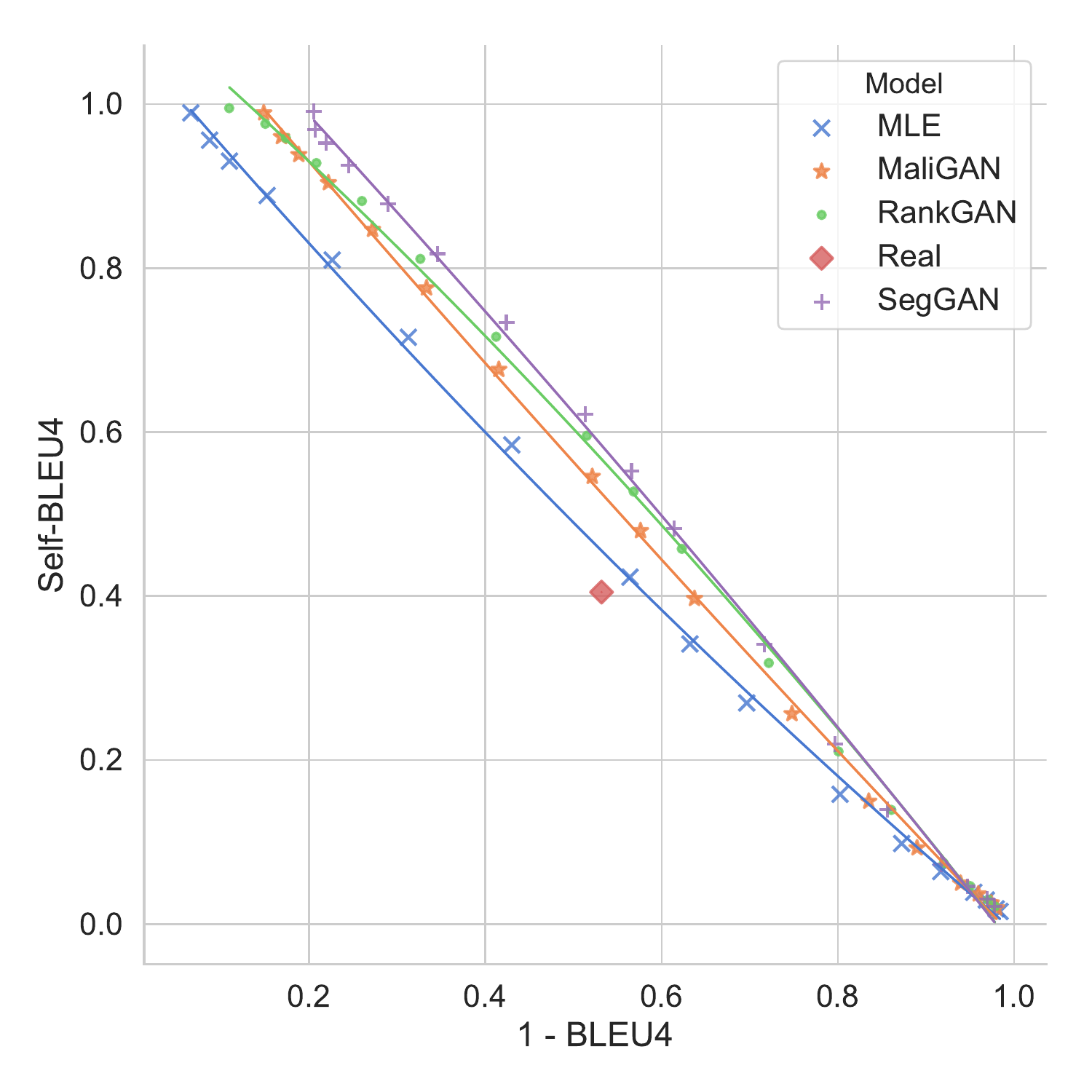}
    	\caption{\textit{IMDB} dataset}
    	\label{M2:imdb}
    \end{subfigure}
    \quad
    \begin{subfigure}[b]{0.40\textwidth}
    	\centering
    	\includegraphics[width=0.93\textwidth]{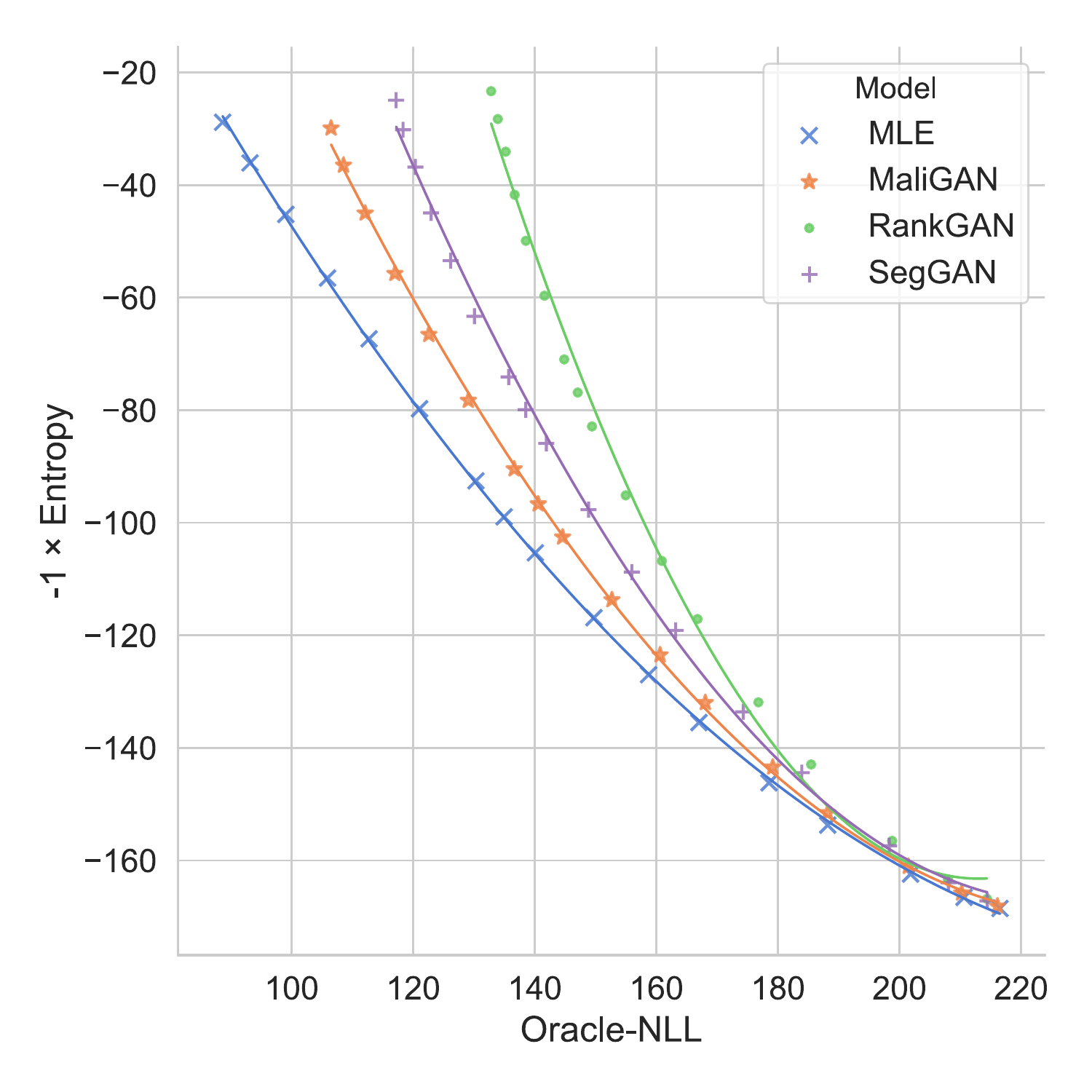}
    	\caption{\textit{Oracle} dataset}
    	\label{M2:oracle}
    \end{subfigure}
    
    \caption{
    Diversity vs. quality measure of various models with temperatures from $1.5^{-3}$ to $1.5^4$ on different datasets. Each point in the plot corresponds to the performance of a model in a special temperature (A second-degree polynomial has been fitted to the points). Lower values in both axes show better ones.
    }
    \label{M2:all}
\end{figure*}
\begin{figure*}[!thb]
    \centering
    \begin{subfigure}[b]{0.3\textwidth}
        \centering
        \includegraphics[trim={0 0 0 0}, clip,width=1\textwidth]{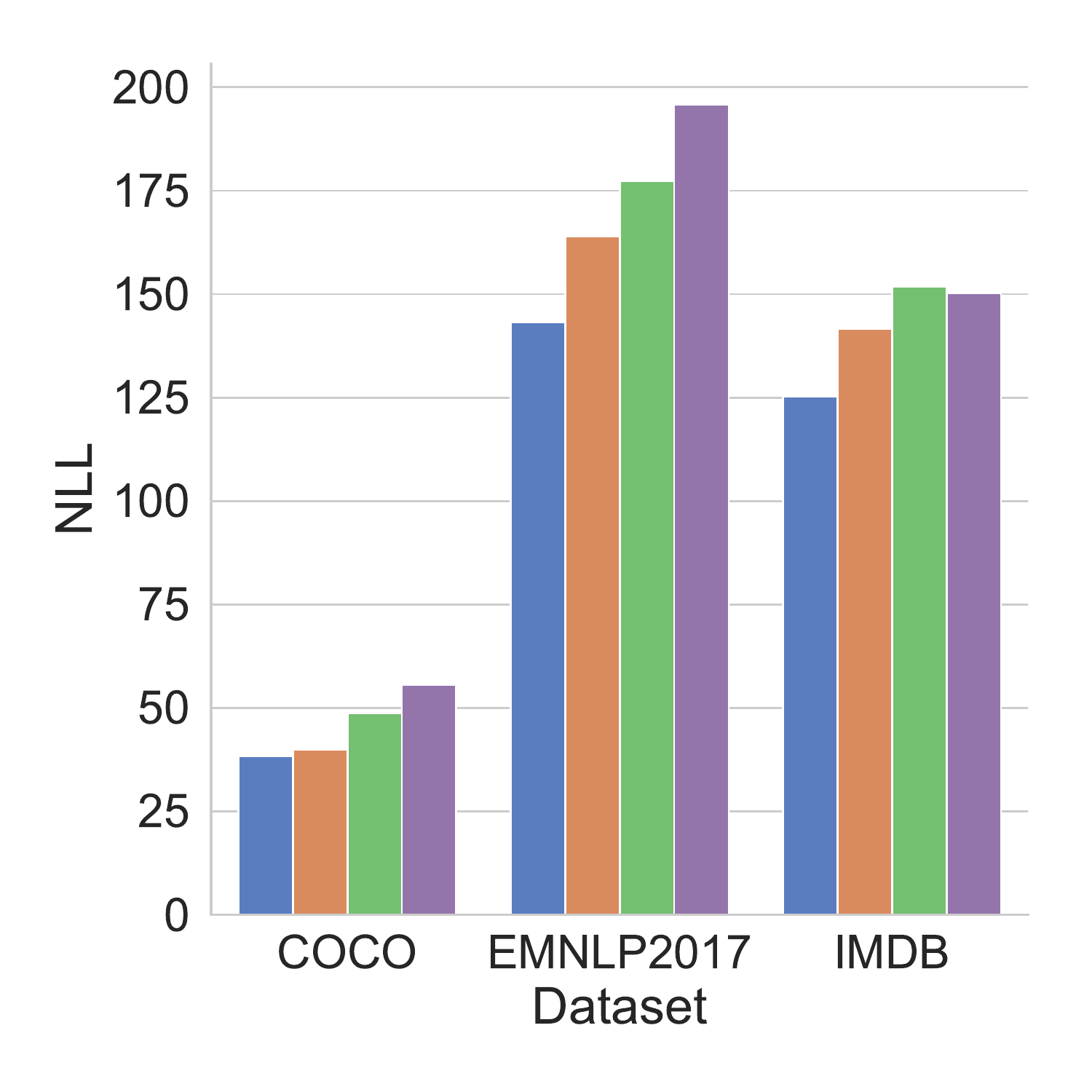}
    	\caption{NLL}
    	\label{M7:Nll}
    \end{subfigure}
    \,
    \begin{subfigure}[b]{0.3\textwidth}
        \centering
        \includegraphics[trim={0 0 0 0}, clip, width=1\textwidth]{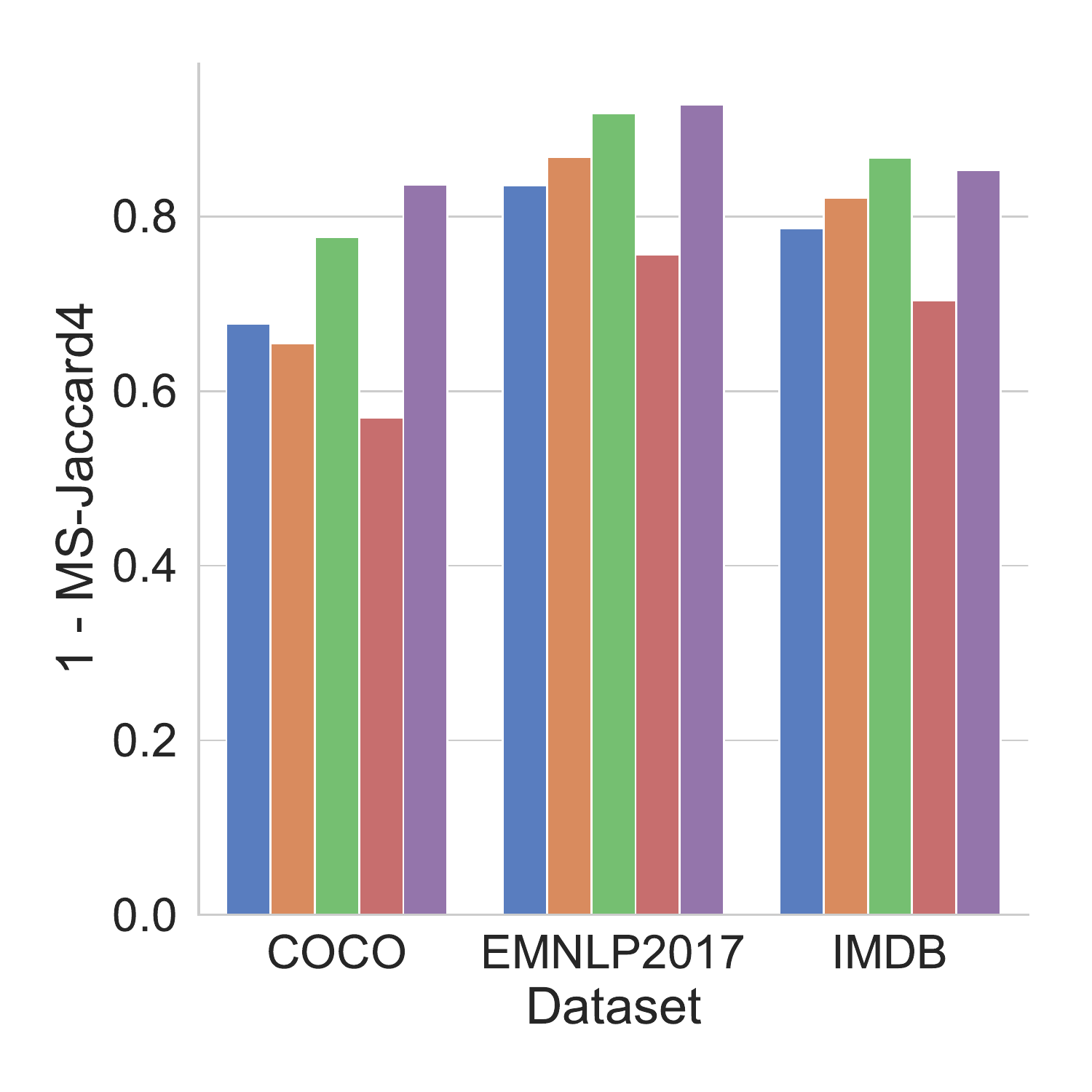}
    	\caption{MS-Jaccard4}
    	\label{M7:Jaccard}
    \end{subfigure}
    \,
    \begin{subfigure}[b]{0.36\textwidth}
    	\centering
        \includegraphics[trim={1.55cm 0 0.6cm 0}, clip, width=1\textwidth]{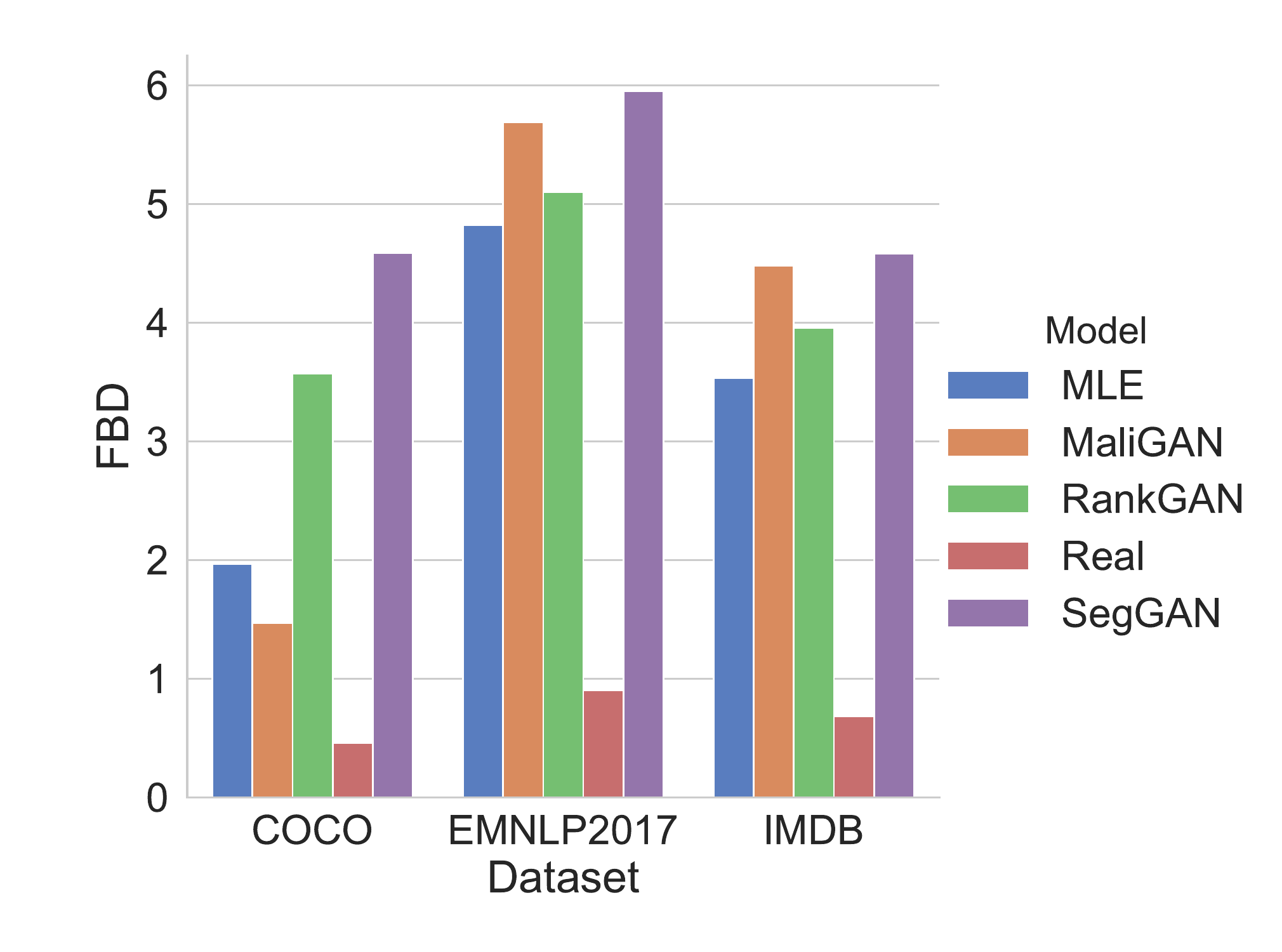}
    	\caption{FBD}
    	\label{M7:FBD}
    \end{subfigure}

    \caption{NLL, $1-$MS-Jaccard4, and FBD scores of all the models without applying temperature (i.e. $T=1$) on different datasets. Lower values show better performance. 
    \label{M7:allreal}
   }
\end{figure*}
\begin{figure}[!htb]
	\centering
	\includegraphics[trim={0.5cm 1cm 0.5cm 0}, clip,width=0.5\textwidth]{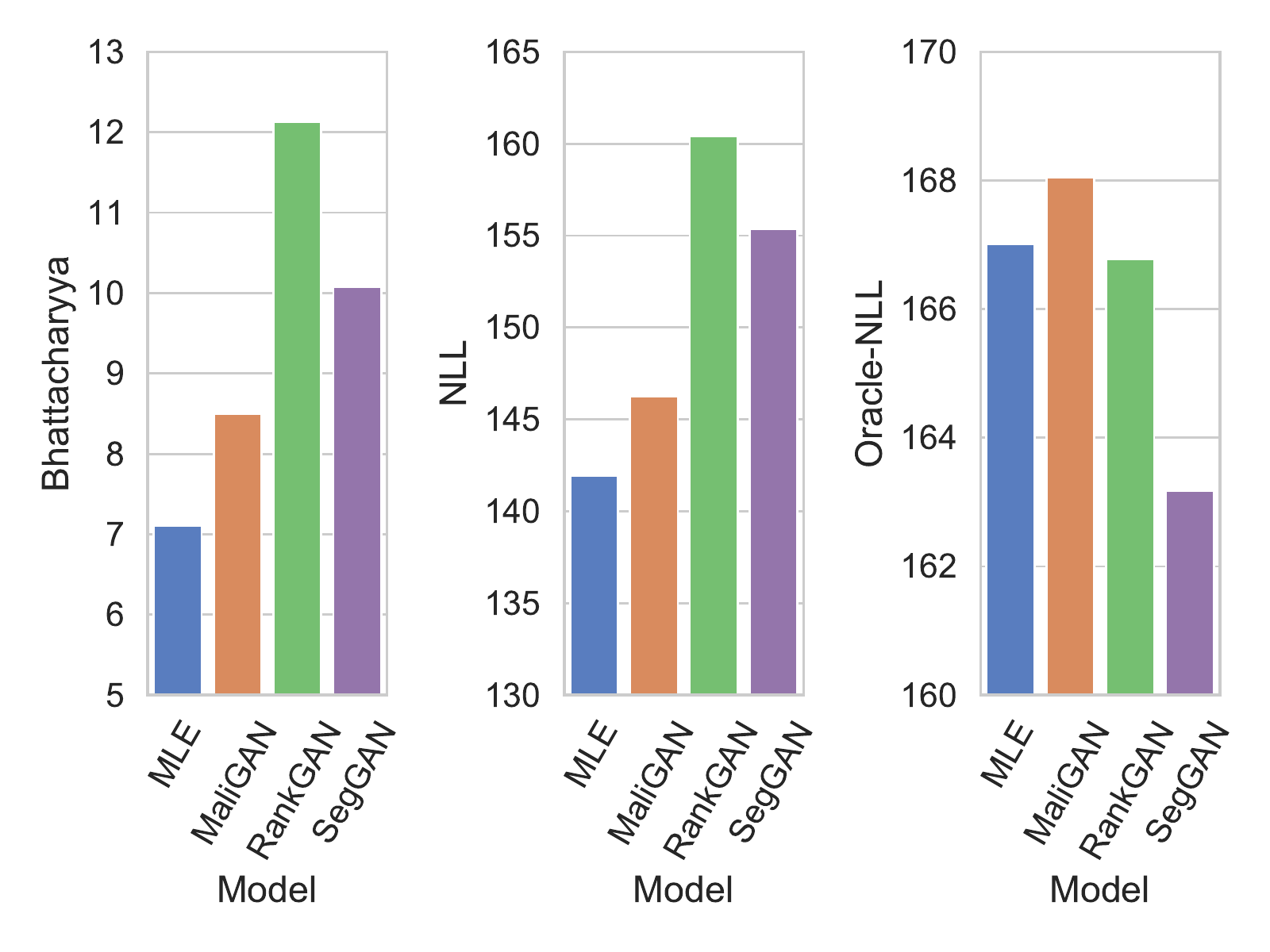}
	\caption{The performance of all models (without applying temperature, i.e. $T=1$) on the Oracle dataset using different measures. Lower values show better performance.}
	\label{M7:Oracle_distance}
\end{figure}
\begin{figure}[!htb]
	\centering
	\includegraphics[trim={0 0 7.0cm 0},clip,width=0.5\textwidth]{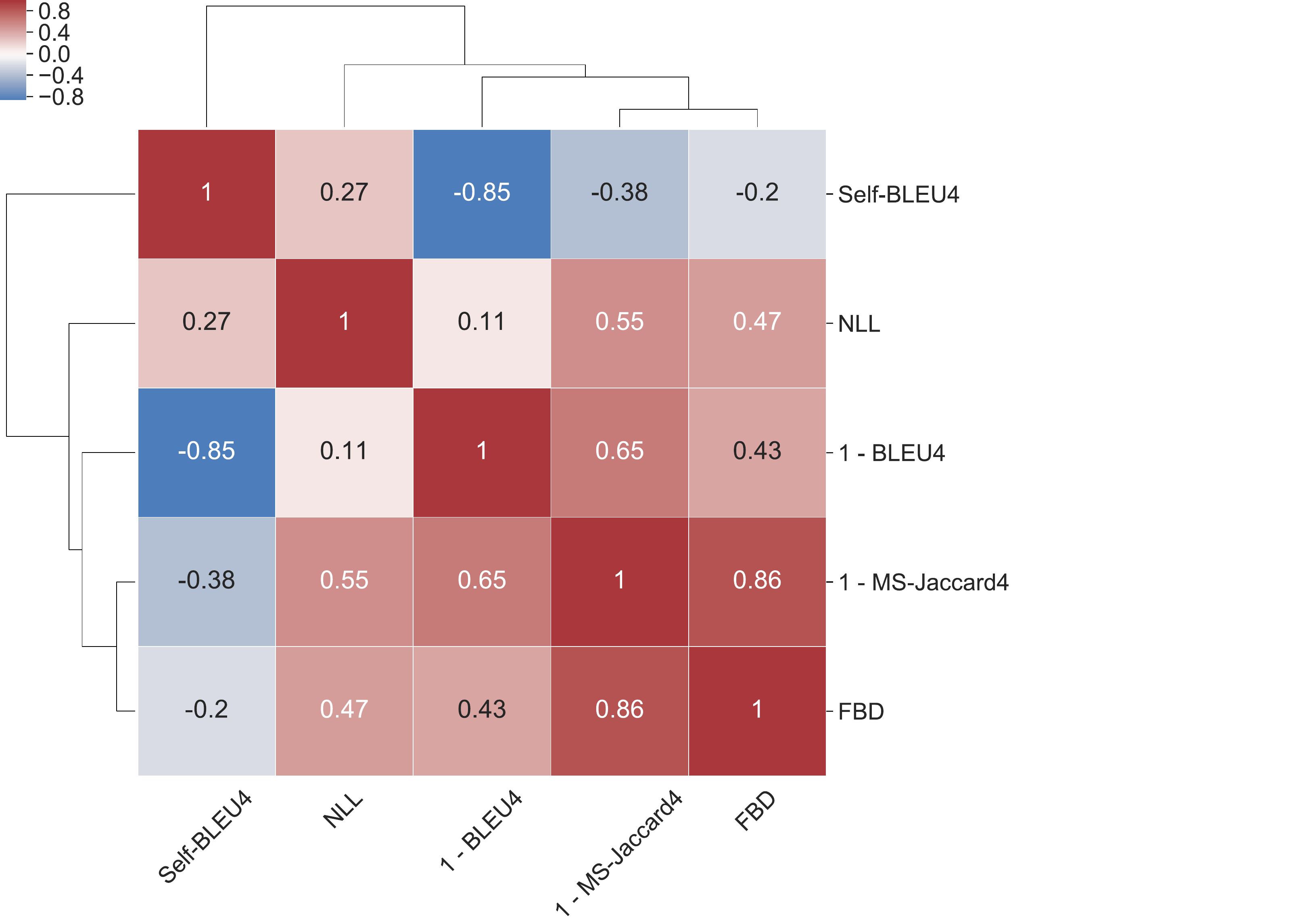}
	\caption{Pearson correlation of all metrics when aggregating results on the real world text datasets and all temperatures.}
	\label{M6:all}
\end{figure}

We intend to show that the proposed metrics are correlated with the analysis of the whole space of quality-diversity obtained by changing the temperature. In fact, using the proposed metrics we can usually predict the behavior of the model in whole spectrum without needing to provide this quality-diversity space.

Fig.~\ref{M2:all} shows the diversity against quality measures with different values of temperature. Figs.~\ref{M2:coco}, \ref{M2:emnlp}, and \ref{M2:imdb} consider Self-BLEU4 as diversity and BLEU4 as quality measure for each of the methods on real-world COCO, EMNLP2017, and IMDB datasets. The metrics are also evaluated on the train data itself which is called Real in the mentioned figures.
Moreover, for Oracle dataset, since we have the probabilistic distribution of data, we can compute the likelihood of the generated samples by the model in the real distribution (i.e. Oracle) to find the quality of the generated samples. Therefore, the Oracle-NLL is used as quality measure of the methods on the synthetic dataset in Fig. \ref{M2:oracle} and Entropy is used as a diversity measure in this figure. 

On the other hand, Figs.~\ref{M7:allreal} and \ref{M7:Oracle_distance} present the performance of different methods (with $T=1$) on non-synthetic and synthetic datasets respectively. 
It is worth noting that NLL, Entropy, and Bhattacharyya of Real could not be computed, since we do not have a model for real data and just considering training data as its samples. 
According to Fig. \ref{M7:Jaccard}, the ordering of the methods obtained by MS-Jaccard4 on these datasets is almost always consistent with the ordering of the methods according to their dominance in Figs.~\ref{M2:coco} to \ref{M2:imdb}. For example, in Fig. \ref{M2:emnlp} that shows results on EMNLP2017 dataset, the best method which dominates others is MLE, the second best is MaliGAN, the third one is RankGAN, and SeqGAN is the last one that under-performs all other methods. Consistently, the proposed MS-Jaccard4 measure shown in Fig. \ref{M7:Jaccard} provides the same ordering. Moreover, the ordering of the methods according to FBD metric in Fig. \ref{M7:FBD} on different datasets is almost always consistent with their ordering obtained by analyzing the whole spectrum in Figs.~\ref{M2:coco} to \ref{M2:imdb}. 
For the oracle dataset \ref{M7:Oracle_distance}, the proposed Bhattacharyya distance of the distributions introduced in Section \ref{measure:oracle} is consistent with the ordering obtained in Fig.~\ref{M2:oracle}.

Finally, we display the Pearson correlation of different metrics on real datasets in Fig.~\ref{M6:all}. According to this figure, the proposed metrics for real-world datasets, i.e. $1-$MS-Jaccard and FBD, are highly correlated. Besides, among the measures, these are the most correlated ones to NLL.

\section{Conclusion}
In this paper, we first discussed shortcomings of the existing measures for evaluating text generation models. Then, we proposed some measures to more effectively specify the capability of models in generating both qualified and diverse texts. The MS-Jaccard as an n-gram based metric was firstly introduced that is capable of measuring both the quality and coverage of methods in text generation. Then, a feature-based metric FBD which is based on the BERT model was introduced. Moreover, for oracle training mode in which the generatorʼs density can also be calculated, we proposed to use (estimation of) divergences like Bhattacharyya defined on probability distributions as a metric to compute the distance of the generative model and the oracle. Finally, the performance of different text generation models were evaluated, the obtained results were analyzed and showed that the proposed metrics have high correlations and are almost consistent with the dominance ordering of models in quality-diversity spectrum.


\bibliography{naaclhlt2019}
\bibliographystyle{acl_natbib}
\end{document}